\documentclass[letterpaper]{article} 
\usepackage[submission]{aaai24}  
\usepackage{tikz}
\usepackage{times}  
\usepackage{helvet}  
\usepackage{courier}  
\usepackage[hyphens]{url}  
\usepackage{graphicx} 
\usepackage{natbib}  
\usepackage{caption} 
\usepackage{hyperref}
\usepackage{graphicx}
\usepackage{booktabs}
\usepackage{subcaption}
\usepackage{setspace} 

\usepackage{algorithm}
\usepackage{algorithmic}

\usepackage{newfloat}
\usepackage{listings}
\DeclareCaptionStyle{ruled}{labelfont=normalfont,labelsep=colon,strut=off} 
\floatstyle{ruled}
\newfloat{listing}{tb}{lst}{}
\floatname{listing}{Listing}
\title{Foundation Models for AI-enabled Biological Design}

\author {
    Asher Moldwin\textsuperscript{\rm 1},
    Amarda Shehu\textsuperscript{\rm 1}
}
\affiliations {
    \textsuperscript{\rm 1}Department of Computer Science, George Mason University, Fairfax, VA, USA\\
    amoldwin@gmu.edu, amarda@gmu.edu
}

\usepackage{bibentry}

\usepackage{multirow}
\usepackage{amssymb,amsmath,amsthm,enumitem}
\usepackage[linewidth=1.2pt,linecolor=black]{mdframed}
\usepackage{algorithm}
\usepackage{algorithmic}

\begin{document}
\maketitle

\begin{abstract} This paper surveys foundation models for AI-enabled biological design, focusing on recent developments in applying large-scale, self-supervised models to tasks such as protein engineering, small molecule design, and genomic sequence design. Though this domain is evolving rapidly, this survey presents and discusses a taxonomy of current models and methods. The focus is on challenges and solutions in adapting these models for biological applications, including biological sequence modeling architectures, controllability in generation, and multi-modal integration. The survey concludes with a discussion of open problems and future directions, offering concrete next-steps to improve the quality of biological sequence generation.
\end{abstract}

\section*{Introduction}
\label{sec:Introduction}

Natural language processing (NLP) has undergone a recent revolution driven by Transformer-based neural networks and the attention mechanism.
These architectures have enabled large language models (LLMs) to achieve remarkable performance on natural language understanding and generation tasks, many of which were once thought to be the exclusive domain of human intelligence. While this transformation in NLP has captured widespread attention, a quieter but equally significant revolution has been unfolding in the biological sciences.

Advances in molecular biology, particularly in large-scale data collection initiatives such as the Protein Structure Initiative\cite{ProteinStructureInitiative}, the Human Genome Project\cite{Collins1995HumanGenome} and structural genomics efforts\cite{dawson2017cath,jones2014structural,bank1971protein}, have paved the way for a new era of biological research. 
The rapid development of next-generation sequencing technologies has led to a wealth of widely available genomics, proteomics, and metabolomics data. 
This multi-omics view of organisms has driven  breakthroughs such as AlphaFold's\cite{jumper2021highly} success in solving protein structure prediction, earning two of its its main authors, Demis Hassabis and John Jumper, the 2024 Nobel Prize in Chemistry. 
Perhaps more importantly, these rapid developments are catalyzing 
a boom in bioinformatics methodologies for various prediction and generation tasks.

The aim of this survey is to provide an overview of recent contributions to AI-enabled biological design, where the goal is to leverage biological data and new methods to design and engineer biological entities with impactful applications in health, drug discovery\cite{blanco2023role}, synthetic biology\cite{voigt2020synthetic}, and material sciences\cite{tang2021materials}. Notably, David Baker, the third recipient of the 2024 Nobel Prize in Chemistry, was recognized for pioneering work in protein design.

While many neural network architectures and machine learning methods are being developed for biological design, this survey focuses on a particularly promising frontier: Foundation Models (FMs). These models, capable of learning general representations from vast datasets in a task-agnostic setting, offer exciting opportunities for biological applications. Our understanding of FMs has evolved in recent years, and we adopt a broad definition that aligns with the Stanford University Human-Centered AI group's understanding\cite{bommasani2021opportunities} in coining the term: FMs are architecture-agnostic and defined by their ability to learn from task-agnostic pre-training, making them applicable across a range of tasks. Although some researchers narrowly define FMs as sequential models or strictly equate them with LLMs, we take the position that FMs should be considered more broadly for their capacity to learn deep representations that can be leveraged for downstream tasks.

Given the rapid progress in FMs, it is timely to survey this expanding landscape.
These models are quickly transforming AI research, with new developments appearing almost weekly, particularly in biological design applications. While this survey cannot be fully comprehensive, it focuses on key areas of concentrated activity or where significant challenges are being articulated and addressed. Specifically, we examine FMs that, leveraging the analogies between natural language and biological sequences, directly support biological sequence-design tasks; though this focus may be somewhat narrow, we believe it encompasses some of the most exciting developments in the field.

The survey highlights sequence-based FMs and focuses on commonly-applied architectures, such as the Transformer, Diffusion, and State Space Model (SSM) architectures. While FMs can encompass a broader range of architectures, we concentrate on these three due to their demonstrated success in recent biological design literature. Moreover, while FMs can support a variety of prediction and downstream tasks, our focus on biological design narrows our attention to generative tasks, where models not only analyze existing biological data but also aim to create novel biological entities with desired properties. 
As we will discuss, this type of biological design includes \textit{de-novo} drug design, protein engineering, and DNA design through gene-editing.

\section{Background and Preliminaries}

\subsection{FMs}

While FMs do not imply a specific neural architecture, certain architectures are more naturally suited to the paradigm of representation learning, which is a critical component in modern FMs. We begin by formalizing the concept of implicit representation learning and illustrate it using a decoder-only text-based Transformer, before introducing the diffusion and SSM-based paradigms.

Given an input sequence $X_{input} = \{x_1,x_2,...,x_T\}$, where each token $x_i$ belongs to a vocabulary $V$, a foundation-model must learn a function to compute an internal representation vector $h_{final}$ such that $h_{final}$ encodes the necessary information to reliably optimize the pre-training objective for different values of $X_{input}$.

In deep learning, this function is typically parameterized such that it can be represented as a series of matrix multiplications and nonlinear transformations.  Each network layer computes an intermediate representation $h_l(x) = \sigma\left(W_l h_{l-1}(x) + b_l\right)$ where $W_l$ is the trainable weight matrix layer $l$, $b_l$ is the bias vector, and $\sigma$ is a nonlinear activation function. The final output layer maps $h_{final}$ to the appropriate form to compute the loss with respect to ground truth.

\subsubsection{Transformer Architecture}
In Transformer architectures\cite{vaswani2017attention}, sequences are handled by maintaining separate representations for each token at each layer, giving: $h_i^{(l)} = \text{TransformerLayer}^{(l)}\left(h_1^{(l-1)}, h_2^{(l-1)}, \dots, h_T^{(l-1)}\right)$. This allows the model to create contextualized representations through the attention mechanism, where each token’s representation at layer $l$ depends on the representations of all tokens from the previous layer. Specifically, for each token embedding $h_i^{(l-1)}$, Transformer attention computes attention scores using the query, key, and value projections: $Q_i = W_q h_i^{(l-1)}$, $K_j = W_k h_j^{(l-1)}$, and $V_j = W_v h_j^{(l-1)}$ for all $j$, where $W_q$, $W_k$, and $W_v$ are trainable weight matrices. The layer representation $h_i^{(l)}$ is then computed as a weighted sum of the values: $h_i^{(l)} = \sum_{j=1}^{T} \text{softmax}(Q_i \cdot K_j) V_j$.

In this formulation, attention incurs a computational cost during training proportional to the number of input tokens squared, rendering its use for long sequences impractical.  Memory usage during inference scales linearly with the sequence length, as key and value representations for all tokens need to be stored, further limiting its efficiency.

To effectively leverage the representational power added by the inclusion of context, the use of general, self-supervised pre-training objectives has proven crucial for learning representations that can be repurposed for other downstream tasks.
For decoder-only models, the pre-training objective is typically autoregressive language modeling, where the model predicts the next token in a sequence given all previous tokens.
Formally, the model learns to approximate the probability distribution $P(x_i | x_1, x_2, \dots, x_{i-1})$.  The attention mechanism is thus constrained to left-to-right context, ensuring that each token attends only to previous tokens: $h_i^{(l)} = \text{Attention}^{(l)}(h_1^{(l-1)}, h_2^{(l-1)}, \dots, h_{i-1}^{(l-1)})$.

Finally, while the attention mechanism handles dependencies between tokens, applying attention multiple times in parallel at each layer (multihead attention), and stacking many Transformer layers in a deep network builds the capacity to learn complex and hierarchical relationships in data.

\subsubsection{State-Space Models (SSMs)}
State-Space Models (SSMs) provide an efficient alternative to Transformers for sequence modeling, particularly for tasks requiring long-range dependencies. While Transformers rely on quadratic self-attention and increasing memory during inference, SSMs achieve subquadratic scaling and constant memory by employing a Recurrent Neural Network architecture and using fixed-size hidden states and linear state-space equations. These equations model state evolution over time as $h_{t+1} = A h_t + B u_{t+1}$, with outputs $y_t = C h_t + \Delta u_t$. Key architectures like S4\cite{gu2021efficiently} enhance SSMs with optimizations for long sequences, while Mamba\cite{gu2023mamba} employs Selective State Spaces to learn dynamic versions the A, B and C matrices whose values depend on the input token.

\subsubsection{Diffusion Modeling}
Models can forgo attention by instead leveraging diffusion processes wherein noise is iteratively introduced to the inputs and then a learned reverse diffusion process is performed to reconstruct realistic inputs. 
This allows global contextual relationships to gradually emerge in the outputs of diffusion models over many iterations of denoising. Formally, diffusion models progressively add random noise to input data over $T$ timesteps, following a Markov process in which each timestep's result depends only on the previous timestep and the added noise. The probability of transforming the input $x_0$ into the noisy state $x_T$ after $T$ steps is given by: 
$q(x_{1:T}|x_0) = \prod_{t=1}^T q(x_t | x_{t-1})$, 
where each transition is defined as: 
$q(x_t | x_{t-1}) = \mathcal{N}(x_t; \sqrt{1-\beta_t} x_{t-1}, \beta_{t} \mathbf{I})$ 
with $\beta_t$ controlling the variance of the added noise at each step. The reverse process, or ``denoising,'' mirrors this forward process, allowing the model to reconstruct the input by removing noise over time. For reverse diffusion, a  neural network $\epsilon_\theta$ is trained to progressively denoise $x_T$ by reversing the diffusion process at each timestep $t$. 
The standard training objective is derived from a variational bound on the data liklihood, giving the loss function $\mathcal{L}(\theta) = \mathbb{E}_{t, x_0, \epsilon} \left[ \lVert \epsilon - \epsilon_\theta(x_t, t) \rVert^2 \right]$.

While diffusion models were originally developed for continuous data, like images, alternative approaches have adapted this framework to discrete data, such as text. In one approach, tokens are mapped into a continuous latent space, where noise is progressively added over $T$ timesteps. The reverse process denoises the embeddings step by step, and a decoder maps the final latent embeddings back to discrete tokens. This allows diffusion models to maintain local coherence while capturing global structure over multiple denoising steps.
Another approach directly applies noise to discrete tokens\cite{li2022diffusion} by either replacing tokens with others sampled from the vocabulary\cite{hoogeboom2021argmax} or by masking them\cite{hoogeboom2021autoregressive}. In the reverse process, a neural network restores the correct token sequence from its noisy version by minimizing the difference between the predicted tokens and the ground truth at each timestep. This iterative denoising process enables diffusion models to generate coherent text, effectively adapting the diffusion framework to discrete data.

\subsection{Biological Design}
Biological design encompasses the creation or modification of biological entities for specific functions or properties. Researchers focus on three primary classes of biological or chemical objects: proteins, DNA/RNA, and small molecules.
Different representations of these objects typically enable different neural network architectures, with the two main representations being sequence-based and graph-based.
Sequence-based representations are derived from the chemical formulas, such as representing proteins as sequences of characters corresponding to the twenty naturally occurring amino acids or representing genomic sequences as strings of the four nucleotide bases. The SMILES representation for small molecules captures atoms as well as bonds and branches. Graph-based representations, on the other hand, model the bonds and interactions (e.g., hydrogen bonds, van der Waals interactions) between atoms or molecular units as edges connecting vertices.

Biological design combines principles from biology, chemistry, and computational science to engineer biological systems. The primary areas of greatest activity are:
(1) Protein engineering: Designing novel proteins or modifying existing ones for enhanced function, stability, or specificity;  (2) Small molecule design: Creating new drug candidates or optimizing existing compounds for improved efficacy and reduced side effects. Advances toward generating biologically functional small molecules or designing novel proteins with precise control over properties—such as binding affinity, solubility, or toxicity—could significantly accelerate therapeutic development by streamlining the identification of drug candidates and (3) Genomic sequence design: Engineering DNA sequences for applications in synthetic biology, gene therapy, or CRISPR-based genome editing.
In this survey we focus on how text-based FMs are adapted for biological design due to their generative capabilities. We will refer to FMs for these three topics as Chemical Language Models (CLMs), Protein Language Models (PLMs), and Genomic Language Models (GLMs).

\section{Taxonomy and Survey}
\begin{figure*}[htbp]
\vspace*{-0.0cm}
\centering
\begin{minipage}{\linewidth}
        \makebox[\linewidth][c]{
            \resizebox{1.0\linewidth}{!}{
\begin{tikzpicture}[
    grow'=right, 
    level 1/.style={sibling distance=23em, level distance=8em,align=center,text width=10em},  
    level 2/.style={sibling distance=4em, level distance=10em,align=center,text width=8em},   
    level 3/.style={sibling distance=4em, level distance=11em,align=center,text width=9em},    
    every node/.style={rounded corners=5pt, thick, inner sep=5pt}, 
    edge from parent/.style={draw,-latex, thick} 
    ]

\node[draw, text width=10em, align=center, text width=6em] {\textbf{FMs for Biological Design}}
    child { node[fill=blue!50, text width =6em] {Architectures}
        child { node[fill=blue!30,font=\small] {Transformer Models}
            child[level distance=20em, sibling distance=3em, inner sep=0pt] { node[fill=blue!20, font=\small, text width =9em] {
            \colorbox{yellow}{\cite{bagal2021molgpt, wang2021multi, mazuz2023molecule, born2023regression}}
            \colorbox[HTML]{ff9194}{\cite{madani2023large, nijkamp2023progen2, ferruz2022protgpt2,born2023regression}}
            }
            edge from parent node[midway,sloped, fill=blue!10,text width = 6em, inner sep=0pt]{Classic} 
            }
            child[level distance=20em, sibling distance=2em] { node[fill=blue!20, font=\small] {
                \colorbox[HTML]{ff9194}{\cite{krishna2024generalized, gruver2024protein}}
                \colorbox[HTML]{80ef80}{\cite{nguyen2024sequence}}
            }
            edge from parent node[midway,sloped, fill=blue!10,text width = 6em, inner sep=0pt]{Specialized} 
            }
        }
        child { node[fill=blue!30,font=\small] {State Space Models}
            child[level distance=20em, sibling distance=2em]  { node[fill=blue!20, font=\small] {
                \colorbox{yellow}{\cite{ozccelik2024chemical}}
                \colorbox[HTML]{ff9194}{\cite{sgarbossa2024protmamba,lal2024reglm,nguyen2024sequence}}
            }
            edge from parent node[midway,sloped, fill=blue!10,text width = 6em, inner sep=0pt]{Classic} 
            }
            child[level distance=20em, sibling distance=2em]  { node[fill=blue!20, font=\small] {~}
            edge from parent node[midway,sloped, fill=blue!10,text width = 6em, inner sep=0pt]{Specialized} 
            }
        }
        child { node[fill=blue!30,font=\small] {Diffusion Models}
            child[level distance=20em, sibling distance=2em]  { node[fill=blue!20, font=\small] {
                \colorbox[HTML]{80ef80}{\cite{li2024discdiff,senan2024dna}}
            }
            edge from parent node[midway,sloped, fill=blue!10,text width = 6em, inner sep=0pt]{Classic} 
            }
            child[level distance=20em, sibling distance=2em]  { node[fill=blue!20, font=\small] {
                \colorbox{yellow}{\cite{peng2024hitting}}
                \colorbox[HTML]{ff9194}{\cite{alamdari2023protein,krishna2024generalized,gruver2024protein,morehead2023towards}}
            }
            edge from parent node[midway,sloped, fill=blue!10,text width = 6em, inner sep=0pt]{Specialized} 
            }
        }
    }
    child { node[fill=green!50, text width = 6em] {Controllability in Generation}
        child { node[fill=green!30,font=\small] {Fine-tuning}
            child { node[fill=green!20, font=\small] {
                \colorbox{yellow}{\cite{ozccelik2024chemical}}
                \colorbox[HTML]{ff9194}{\cite{madani2023large, nijkamp2023progen2, sgarbossa2024protmamba,krishna2024generalized}}
                \colorbox[HTML]{80ef80}{\cite{nguyen2024sequence,lal2024reglm}}
            } }
        }
        child { node[fill=green!30,font=\small] {Conditional Generation}
            child { node[fill=green!20, font=\small] {
                \colorbox{yellow}{\cite{bagal2021molgpt,wang2021multi, born2023regression}}
                \colorbox[HTML]{ff9194}{\cite{born2023regression,madani2023large, nijkamp2023progen2, sgarbossa2024protmamba,krishna2024generalized}}
                \colorbox[HTML]{80ef80}{\cite{nguyen2024sequence,lal2024reglm,li2024discdiff,senan2024dna}}
            } }
        }
        child { node[fill=green!30,font=\small] {Reinforcement Learning}
            child { node[fill=green!20, font=\small] {
            \colorbox{yellow}{\cite{mazuz2023molecule,wang2021multi}}
            \colorbox[HTML]{ff9194}{\cite{angermueller2019model}}
            } }
        }
        child { node[fill=green!30,font=\small] {Custom Objective Functions}
            child { node[fill=green!20, font=\small] {
            \colorbox[HTML]{ff9194}{\cite{gruver2024protein, born2023regression}}
            } }
        }
        child { node[fill=green!30,font=\small] {Post-Generation Filtering}
            child { node[fill=green!20, font=\small] {
                \colorbox{yellow}{\cite{ozccelik2024chemical}}
                \colorbox[HTML]{80ef80}{\cite{lal2024reglm}}
            } }
        }
        child { node[fill=green!30,font=\small] {Numerical Property Optimization Strategies}
            child { node[fill=green!20, font=\small] {
            \colorbox{yellow}{\cite{born2023regression}}
            \colorbox[HTML]{ff9194}{\cite{born2023regression}}
            } }
        }
        child[sibling distance=4.4em] { node[fill=green!30,font=\small] {Multi-condition Generation \\ and Tradeoff Handling}
            child { node[fill=green!20, font=\small] {
                \colorbox{yellow}{\cite{peng2024hitting,wang2021multi}}
                \colorbox[HTML]{ff9194}{\cite{gruver2024protein}}
            } }
        }
        child[sibling distance=4.3em] { node[fill=green!30,font=\small] {Sampling Algorithms}
            child { node[fill=green!20, font=\small] {
                \colorbox[HTML]{ff9194}{\cite{ferruz2022protgpt2}}
            } }
        }
    }
    child { node[fill=gray!50, text width = 6em] {Multi-Modal FMs}
        child { node[fill=gray!30,font=\small] {Combining Multiple \\ Forms of Biological Sequences}
            child { node[fill=gray!20, font=\small] {
                \colorbox[HTML]{ff9194}{\cite{krishna2024generalized,morehead2023towards}}
            } }
        }
        child { node[fill=gray!30,font=\small,font=\small] {Combining Sequence and \\ Structure Information}
            child { node[fill=gray!20, font=\small] {
                \colorbox[HTML]{ff9194}{\cite{krishna2024generalized,morehead2023towards, madani2023large, sgarbossa2024protmamba}}
            } }
        }
        child { node[fill=gray!30,font=\small] {Incorporating Natural \\ Language and Domain Knowledge}
            child { node[fill=gray!20, font=\small] {
                \colorbox{yellow}{\cite{livne2024nach0}}
                \colorbox[HTML]{ff9194}{\cite{zhuang2024instructbiomol}}
                \colorbox[HTML]{80ef80}{\cite{richard2024chatnt}}
            } 
                child[edge from parent/.style={draw=none}] { node[draw, fill=white, font=\small, text width=9em,align=left] {
                \colorbox{white}{\makebox[0.2cm][c]{\strut}}~~\textbf{Citation Key}
                \colorbox{yellow}{\makebox[0.5cm][c]{\strut}}~~Small Molecules
                \colorbox[HTML]{ff9194}{\makebox[0.5cm][c]{\strut}}~~Proteins
                \colorbox[HTML]{80ef80}{\makebox[0.5cm][c]{\strut}}~~DNA
                }
                }
        }
    }
    };
\end{tikzpicture}
            };
        }
\caption{Taxonomy of recent contributions in FMs for Biological Design. Leaf nodes contain references to the relevant papers, color coded by the type of biological sequence that they discuss.}
\label{fig:taxonomy}
\end{minipage}
\end{figure*}
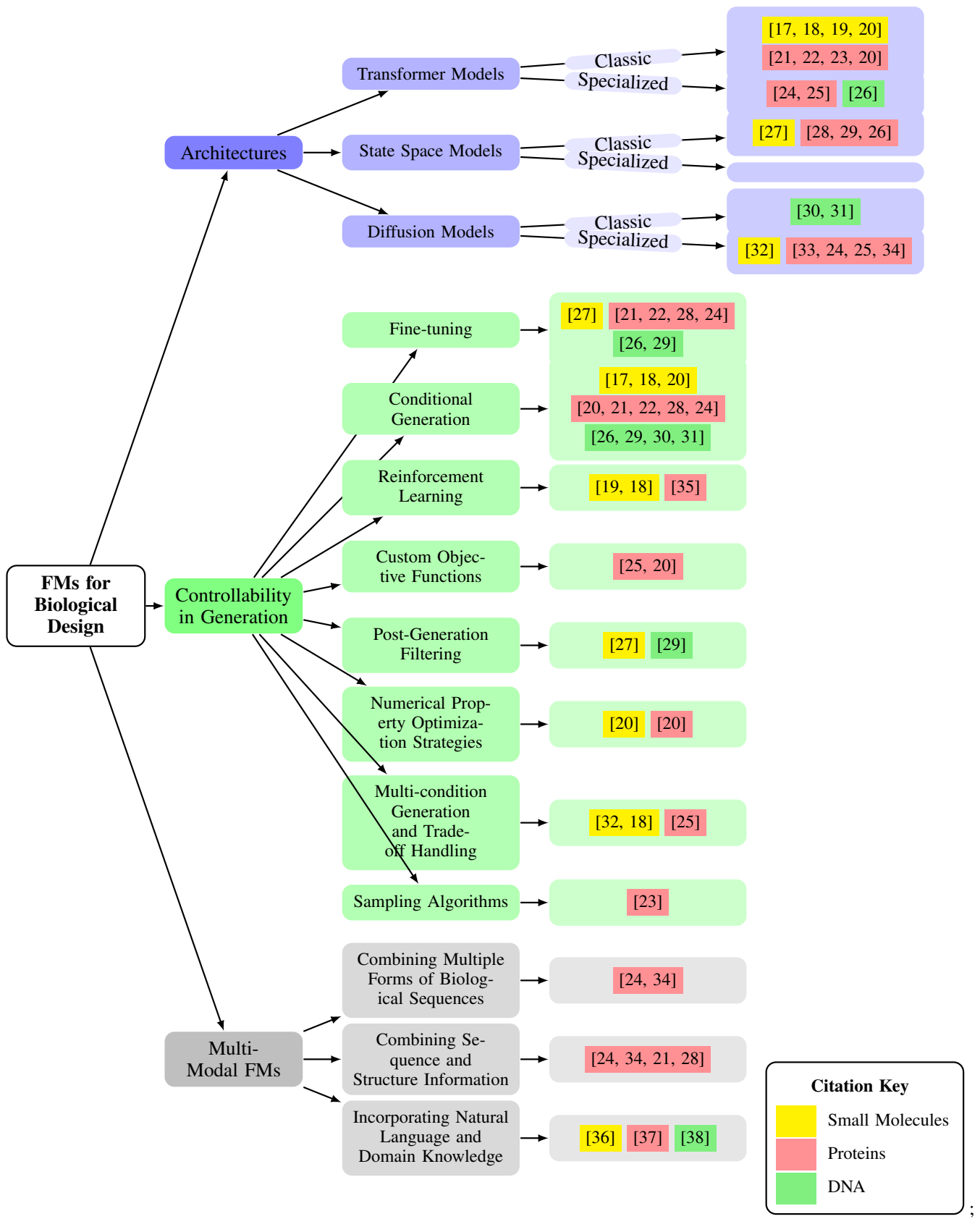

Figure~\ref{fig:taxonomy} shows our taxonomy highlighting recent areas of focus in FMs for Biological Sequence Design. We include references to papers that address each topic in the leaf nodes of the taxonomy tree, color-coded by the biological domain that they discuss. 

Methods aligned with these categories are summarized in Table~\ref{tab:tab1}.

\begin{table*}[htbp]
\centering
\footnotesize
\caption{Summary of current FM models for Biological Design.}
\label{tab:tab1}
\resizebox{0.96\textwidth}{!}{
\begin{tabular}{|p{3cm}|p{1cm}|p{3cm}|p{4cm}|p{4cm}|}
\hline
\textbf{Model Name} & \textbf{Domain} & \textbf{Architecture} & \textbf{Design Goal} & \textbf{Generation Method} \\ \hline
MolGPT \cite{bagal2021molgpt} & Small Molecule & Decoder-only Transformer & Control TPSA, QED, SAS, LogP and molecular scaffolds & Conditional Autoregressive Generation, pre-trained with prepended property condition embeddings \\ \hline
MCMG \cite{wang2021multi} & Small Molecule & Decoder-only Transformer Distilled into RNN & Control Bioactivity, QED, SAS & Same as above, with added reinforcement learning rewarding low property error \\ \hline
Taiga \cite{mazuz2023molecule} & Small Molecule & Decoder-only Transformer & QED, inhibitory potency (plC50) & Policy gradient reinforcement learning rewarding high property values \\ \hline
S4 CLM \cite{ozccelik2024chemical} & Small Molecule & S4-based SSM & MAPk1 kinase inhibitors & fine-tuning for transfer to desired class \\ \hline
DiffuMol \cite{peng2024hitting} & Small Molecule & Diffusion on embeddings, Transformer decoder for denoising & LogP, QED, TPSA, SAS, and scaffolds & Noiseless conditional token anchors  \\ \hline
Regression Transformer\cite{born2023regression} & Small Molecule and Protein & Decoder-only Transformer (XLNet) & Control QED, solubility, lipophilicity for small molecules, fluorescence, stability and Bowan index for proteins & Alternating training scheme that switches between property prediction and conditional sequence generation \\ \hline
Progen \cite{madani2023large} & Protein & Decoder-only Transformer & Proteins associated with specific families, biological functions & Conditional autoregressive generation and fine-tuning \\ \hline
ProGen2 \cite{nijkamp2023progen2} & Protein & Decoder-only Transformer with RoPE\cite{su2024roformer} & Structurally valid proteins, specific folds, antibodies & Fine-tuning, and three-residue motif prompts for antibodies \\ \hline
ProtGPT2 \cite{ferruz2022protgpt2} & Protein & Decoder-only Transformer & Novel proteins that are also plausible, mostly globular proteins & Autoregressive sampling with modified sampling schemes \\ \hline
ProtMamba \cite{sgarbossa2024protmamba} & Protein & Mamba-based SSM & Homologous proteins, valid inpainting-based protein modifications & Autoregressive with homologs for context, or inpainting using Fill-in-the-Middle \\ \hline
EvoDiff \cite{alamdari2023protein} & Protein & Diffusion with Dilated CNN & Proteins from specific families, inpainting functional domains, generating around motifs & MSA-conditioned for family-based, motif-based conditioning and inpainting for scaffolds  \\ \hline
NOS \cite{gruver2024protein} & Protein & BERT-based Encoder-Decoder & Antibodies with high expression yield and binding affinity & Diffusion, multi property value function, Latent Multi-Objective Bayesian Optimization  \\ \hline
RFDiffusionAA \cite{krishna2024generalized} & Protein complexes & Specialized Multi-track attention based architecture & Protein binders for small molecules, including nucleic acids, proteins, and ligand complexes & Condition on ligand as fixed noiseless anchor \\ \hline
MMDIFF \cite{morehead2023towards} & DNA and Protein & One-hot vector for discrete sequences, FrameDiff architecture for 3d structure & Macromolecular Complexes & Joint reverse diffusion with separate loss components for sequence and structure \\ \hline
regLM \cite{lal2024reglm} & DNA & Based on Hyena-DNA\cite{nguyen2024hyenadna}, SSM-like & Cis-regulatory elements with specified levels of activity or cell-type specificity & Special condition tokens for activity level or cell type, outputs filtered using regression model \\ \hline
Evo \cite{nguyen2024sequence} & DNA & StripedHyena using attention and SSM-like blocks & Nucleotide sequences for CRISPR systems, function-conserving transposable elements & Fine-tuning to enable conditioning on special tokens \\ \hline
DiscDiff \cite{li2024discdiff} & DNA & 2-stage VAE, with U-net based denoising networks & DNA for specific species, gene expression levels, or cell types & Conditional generation with absorb-escape algorithm \\ \hline
DNA-diffusion \cite{senan2024dna} & DNA & U-net backbone & Regulatory sequences to control chromatin accessibility & Conditioning on cell-type labels \\ \hline
\end{tabular}
}
\end{table*}

\subsection{Architectures}

\subsection*{Transformer Models}

\textbf{Classic Transformer Models}\\
The SMILES representation facilitates using standard Transformer architectures like GPT and XLNet \cite{yang2019xlnet} for small-molecule design, often with minimal modifications and NLP-optimized hyperparameters. MolGPT \cite{bagal2021molgpt}, MCMG \cite{wang2021multi}, Regression Transformer \cite{moret2020generative}, and Taiga \cite{mazuz2023molecule} employ GPT-like architectures but differ in training and controllability strategies. PLMs like Progen \cite{madani2023large}, Progen2 \cite{nijkamp2023progen2}, and ProtGPT2 \cite{ferruz2022protgpt2} also use standard architectures with minor enhancements. For instance, Progen2 incorporates Rotary Position Embeddings (RoPE \cite{su2024roformer}) and improved parallelization. In contrast, most DNA FMs adopt highly customized attention mechanisms or alternative architectures.
\\
\textbf{Specialized for Biological Design}\\
Transformer-based models often adapt attention mechanisms to suit biological tasks. Examples include RFDiffusionAllAtom \cite{krishna2024generalized}, NOS \cite{gruver2024protein}, and Evo \cite{nguyen2024sequence}. RFDiffusionAllAtom extends RoseTTAFold2’s multi-track attention with atomic-level details for protein-molecule complex design. NOS employs an encoder-decoder Transformer for forward and backward diffusion, optimized for antibody design. Evo uses a hybrid StripedHyena architecture, combining RoPE-based attention with convolutional layers \cite{poli2023hyena}, enabling efficient processing of long genomic sequences (up to 131 kilobases) for tasks like CRISPR DNA design.

\subsection*{State Space Models}

\textbf{Classic SSMs}\\
Many SSM architectures struggle to effectively capture relevant information between tokens separated by large distances in the input. This challenge is significant for protein data, where distant amino acids in the sequence may interact through hydrogen bonds, disulfide bridges, or hydrophobic interactions. Likewise, genomic data contain regulatory elements located hundreds of thousands of tokens away from the genes they regulate. A similar  challenge exists in modeling ring closure and branching in small molecules \cite{ozccelik2024chemical}, though these sequences are considerably shorter.

Several studies have applied Mamba and S4 architectures directly to biological tasks and report superior performance. However, what qualifies as a fair comparison remains unclear (e.g., should models be compared based on the same number of parameters, training time, or inference time?). The CLM introduced by Ozccelik et al. \cite{ozccelik2024chemical} was the first to apply S4 models to SMILES chemical sequences, benchmarking them against other CLMs such as LSTMs and GPT-type models for \emph{de-novo} molecule generation. They demonstrate SSM advantages in generating valid, unique, and novel molecules. 

Various SSM-based protein sequence models have emerged, including ProtMAMBA\cite{sgarbossa2024protmamba}, ProteinMamba \cite{xu2024protein}, and PTMMamba \cite{peng2024ptm}. ProtMAMBA can autoregressively generate new protein sequences, with or without conditional information from homologous sequences. Motivated by long-range dependencies in proteins, it employs a variant of the "fill-in-the-middle" pre-training objective \cite{bavarian2022efficient}, which trains the model to predict masked segments of a sequence using both preceding and succeeding amino acids. This enables more accurate sequence inpainting and generation tasks designed to capture long-range dependencies. ProtMAMBAFoundation is proposed for general tasks, such as sequence generation and inpainting, while ProtMAMBAFine-Tuned is designed specifically for inpainting.

While Mamba and S4 models are well-established in the State-Space Modeling paradigm, the precise definition of SSMs can be ambiguous. Following Benegas et al. \cite{benegas2024genomic}, models based on the Hyena architecture\cite{poli2023hyena} can also be classified as SSMs due to shared features like data-dependent gating and long convolutions. Hyena, which generalizes techniques from H3\cite{fu2022hungry}, excels in long-range sequence comparisons, making it highly effective for DNA modeling. In the realm of DNA-sequence generation, models like HyenaDNA \cite{nguyen2024hyenadna} serve as backbones. This efficient long-sequence architecture is critical, as HyenaDNA is trained on sequences of up to 1 million nucleotides. It allows the model to capture long-range interactions at single-nucleotide resolution, which is crucial for tasks like detecting regulatory elements, mutations, and other genomic features spanning large distances. RegLM\cite{lal2024reglm} is a generative model that uses a Hyena-based backbone, and Evo \cite{nguyen2024sequence} relies on a variant called StripedHyena, combining both attention and Hyena Layers. 

\subsection*{Diffusion Models}

\textbf{Classic Diffusion Models}\\
Diffusion models have become popular for generating continuous data, such as images or molecular graphs. 
Recently, these models have also been extended to biological sequences by embedding discrete token inputs. For example, DNA-Diffusion \cite{senan2024dna} generates regulatory sequences controlling chromatin accessibility and gene expression. It uses a U-net convolutional architecture, similar to image diffusion networks like DALL-E \cite{ramesh2021zero}. By applying a transformation similar to one-hot encoding—where each nucleotide is represented within a continuous range of [-1,1]—DNA-Diffusion is able to introduce Gaussian noise to discrete biological sequences, facilitating the diffusion process. 
Building on similar principles, DiscDiff \cite{li2024discdiff} leverages a U-net with ResNet blocks for DNA generation, using a Variational Autoencoder to encode nucleotide sequences into a continuous latent space for diffusion, before decoding them back into discrete form after the reverse diffusion process.\\
\textbf{Specialized for Biological Design}\\
EvoDiff \cite{alamdari2023protein} explores two distinct forward diffusion processes tailored for protein modeling:
Order Agnostic Autoregressive Diffusion (OADM), which involves randomly masking tokens, and the Discrete Denoising Diffusion Probabilistic Model (D3PM), which applies mutation-based noise by replacing amino acids according to a transition matrix derived from natural mutation probabilities. This mutation-driven approach allows EvoDiff to navigate biologically meaningful regions of evolutionary space, increasing the likelihood that the generated proteins will retain essential properties such as functionality, stability, and activity—features typically conserved in naturally-evolved proteins.

We observe hybridizations of classic diffusion and transformer architectures to capture broader biological contexts. NOS\cite{gruver2024protein} adapts a BERT-small model  encoder-decoder model to better integrate contextual information in protein sequences during diffusion, while DiffuMol \cite{peng2024hitting} targets small-molecule design, incorporating a Transformer decoder to generate molecules with desired properties like QED and LogP.

In multi-modal data contexts, RFDiffusionAA extends RoseTTAFold All-Atom for diffusion-based generation of protein-molecule complexes, utilizing both sequence and structure. Similarly, MMDIFF \cite{morehead2023towards} introduces joint diffusion across DNA and protein sequences, incorporating distinct loss components for sequence and structural modalities to enable coordinated generation of biologically coherent macromolecular complexes.

\subsection{Controllability in Generation}

A key challenge in advancing FMs for biological design is achieving fine-grained control over generated data. We note that biological design is inherently an engineering endeavor. The designed entities are intended to be ultimately synthesized in wet laboratories and then operationalized for particular outcomes. So, the issue of control is inherent to successful, synthetically-accessible and operationalizable design.


For small molecules, continuous numerical properties may need to be controlled, including Drug-likeness, LogP, Molecular Weight, Synthetic Accessibility, Toxicity, and Topological Surface Area. One may also require particular classes of molecular compounds, such as ``kinase inhibitors'' or ``Quaternary Ammonium Compounds'' depending on the downstream task/application. For proteins, one typically controls for specific functions or similarity to other known proteins. For DNA, it may be necessary to ensure specific regulatory properties, such as promoters or enhancers, or design guide RNAs (gRNAs) to facilitate CRISPR-based gene editing.

\subsection*{Fine-Tuning}
Many of the models discussed above rely on some form of fine-tuning to adapt a pre-trained FM for controlled generation. In its simplest form, this involves continued training with the same pre-training objective but on a smaller subset of data that matches the desired properties/criteria.
Ozccelik et al\cite{ozccelik2024chemical} pre-train- their S4-based CLM on 1.9 million SMILES, then fine-tune it on 68 manually-annotated molecules known to inhibit MAPK1, generating new candidates likely to have high binding affinity to MAPK1. Madani et al. \cite{madani2023large} fine-tune the Progen model on lysozyme families after pre-training on 280 million protein sequences, then generating proteins with catalytic efficiencies comparable to natural lysozymes despite low sequence identity. 

Fine-tuning can be computationally expensive, especially for large models like ProGen, with 1.2 billion parameters, or insufficient when more granular control is required. Nevertheless, fine-tuning is often a necessary step before applying more controlled generation techniques, such as conditional generation. It is worth noting that fine tuning may not be effective if the dataset with properties of interest is particularly small relative to the pre-training dataset.

\subsection*{Conditional Generation}

Conditional generation offers another method for controlling model outputs by embedding ``prompts'' or ``control tags'' into the input, which guide the generation toward desired characteristics. RegLM \cite{lal2024reglm} employs conditional generation to design synthetic cis-regulatory elements ( CREs-e.g., promoters, enhancers), where autoregressive generation is conditioned on a starter fragment that guides the prediction of subsequent tokens. Similarly, Evo \cite{nguyen2024sequence} fine-tune a model on $82,430$ loci containing CRISPR-Cas sequences, using prompt tokens (e.g., ``cas9,'' ``cas12,'' ``cas13'') to optimize gRNA sequences for improved efficiency and reduced off-target effects.

Progen2 conditions the generation of antibody (protein) sequences on three-residue motif prompts, while ProtMamba enables homologous protein generation by conditioning on proteins from the desired family as context. RFDiffusionAA \cite{krishna2024generalized} ensures accurate protein-ligand complex generation by conditioning the denoising process on the presence of an unaltered ligand during diffusion.
In contrast, MolGPT~\cite{bagal2021molgpt} embeds control tokens directly in the pre-training inputs. 
While effective, this approach retrains the model from scratch for each new property or combination of properties, diverging from the more general FM paradigm.

While conditional generation enables control over fine-grained and numerical properties, it presents several limitations. Chief among these is the reliance on examples in the training or fine-tuning data, which can constrain the model's ability to discover novel ways of achieving desired properties. Additionally, the model may develop a bias toward retaining features correlated with the target properties from the training data, potentially hampering diversity in generated data.

\subsection*{Reinforcement Learning}

Reinforcement learning (RL) offers a promising alternative by guiding the generation with a reward function. In RL, an agent learns to make decisions through feedback from an environment, optimizing for specific objectives. In sequence generation, each token is treated as an action, and the model is fine-tuned based on how well the generated sequence meets the desired conditions. RL’s flexibility in handling complex, non-linear reward landscapes makes it ideal for multi-objective optimization, where properties like bioactivity and molecular diversity must be balanced.

The MCMG model~\cite{wang2021multi} employs RL to control multiple properties in small-molecule generation. Initially, a conditional Transformer generates molecules based on property embeddings. To improve the model's efficiency and balance diversity and novelty, the trained model is distilled into a simpler recurrent model, which is then fine-tuned with RL. During RL training, the model generates molecules autoregressively, with a reward function evaluating properties such as bioactivity, drug-likeness, and synthetic accessibility. Invalid molecules receive a reward of zero, and the reward is incorporated into an augmented likelihood loss function that balances property optimization with sequence validity.

Similarly, Mazuz et al.~\cite{mazuz2023molecule} combine a Transformer with RL, using a policy gradient approach and the REINFORCE algorithm \cite{williams1992simple}. After pre-training on SMILES strings, the model is fine-tuned to optimize molecular properties like QED and inhibitory concentrations (pIC50). A reward function proportional to the property values is applied for fully generated, valid molecules, while intermediate generation steps receive zero reward. The model optimizes the expected return, with a high discount factor ($\gamma = 0.99$) to prioritize near-term rewards during sequence generation. Despite these successes, we note that RL is often impractical due to the challenges of designing effective reward functions, which demand significant biological and computational insight, especially given the high-dimensional action spaces in biological design.

\subsection*{Custom Objective Functions}
Non-RL methods also employ custom loss functions to directly evaluate sequence properties, allowing for property-specific optimization during generation. NOS\cite{gruver2024protein} uses custom objective functions to optimize properties such as binding affinity and expression yield. This is achieved by incorporating a value function, $v_\theta (w)$, trained via a separate discriminator (or value network), which shares hidden layers with the generative diffusion model up to a certain depth. The discriminator is trained to predict the target objective for proteins in the training set.
Once trained, the value function is applied to the hidden states of the sequence representation in the denoising network, and the model uses gradient ascent on these hidden states to maximize the objective value.

The Regression Transformer takes a similar approach by alternating between sequence generation and numerical-property regression tasks. It introduces a custom ``property prediction objective" that masks numerical tokens representing property values, requiring the model to predict these values from the associated textual sequence:
$J_P = \max_\theta \mathbb{E}_{z \sim Z_T^p} \left[ \log p_\theta(x^p | x^t) \right]$,
where $x^p$ are the property tokens, $x^t$ are the associated textual tokens, and 
$z\sim Z^p_T$
denotes the constrained permutation order ensuring that only property tokens are masked.
Additionally, a ``self-consistency loss" is introduced to ensure the generated sequence adheres to the primed property by using the model's own property prediction as a target during generation:
$J_{SC} = J_G(x)+\alpha \cdot J_p (\hat{x})$,
where $J_G$ is the conditional text generation objective, $J_p$ is the property prediction objective. By alternating between these custom loss functions, the model can effectively optimize conditional generation for fine-grained numerical properties.

\subsection*{Post-Generation Filtering}
A key measure of success for FMs in biological sequence design is the ability to reliably generate sequences worthwhile for practical validation. If \textit{in-silico} validation is cost-effective, high-throughput filtering can screen large numbers of candidates, discarding those that fail to meet desired conditions before advancing to \textit{in-vitro} validation.

An example of this is the SSM-based CLM proposed by Ozccelik et al\cite{ozccelik2024chemical}, where, after fine-tuning on  MAPK1 inhibitors as mentioned above, they next generated 256,000 molecules. The molecules are ranked based on log-likelihood scores, computed by subtracting pre-training scores from fine-tuning scores to focus on newly learned bioactivity features. From the top 5,000 molecules, scaffold similarity filtering divided them into high and low similarity groups. The highest-scoring molecules from each group were further evaluated using molecular dynamics simulations to assess binding affinity.

Similarly, RegLM \cite{lal2024reglm} apply post-generation filtering to synthetic CREs by using a sequence-to-function regression model to predict activity levels. Sequences outside two standard deviations from the mean activity were discarded. The model also filtered sequences for cell type specificity, selecting those with the highest specificity based on predicted activity in target and off-target cell lines. Biological realism was ensured through metrics like GC content and k-mer frequency, ensuring the synthetic sequences mimicked natural CREs.

\subsection*{Numerical Property Optimization Strategies}
In biological design, many properties of interest are numerical, such as synthetic accessibility, drug-likeness, lipophilicity, and molecular weight. These molecular properties, evaluated using tools like RDKit, were considered in models like MolGPT\cite{bagal2021molgpt} and DiffuMol\cite{peng2024hitting}. Additional numerical properties include bioactivity\cite{wang2021multi}, solubility\cite{born2023regression}, and inhibitory potency\cite{mazuz2023molecule}. For proteins, numerical properties like interaction potential (e.g., the Boman index), fluorescence, and stability (from datasets like TAPE \cite{rao2019evaluating}) are of interest \cite{born2023regression}. In DNA design, properties like gene expression levels are key, as seen in DiscDiff \cite{li2024discdiff}, which optimizes regulatory elements to control expression.

Several models incorporate these numerical targets directly into training. For instance, RL-based methods \cite{mazuz2023molecule, wang2021multi} optimize numerical properties as part of their reward functions. One specialized approach is the Regression Transformer \cite{born2023regression}, which encodes numerical properties as tokens within a decoder-only architecture. To handle numerical tokenization, the authors introduced a tokenization scheme for floating-point numbers and developed numerical encodings, similar to positional encodings, to help the model understand numerical relationships.
MolGPT \cite{bagal2021molgpt}, by contrast, reserves specific dimensions in the input embeddings for numerical values, avoiding explicit tokenization. This differs from the common NLP approach used in models like T5 \cite{raffel2020exploring} and GPTs \cite{brown2020language}, which apply word-piece tokenization to numbers. 

\subsection*{Multi-Condition Generation and Tradeoff Handling}

Generating molecules with multiple specified properties can be challenging, especially when the desired combinations are rare in the training data or constrained by physical and chemical limitations. A key challenge is managing tradeoffs, where the model must balance generating valid and diverse molecules that meet all specified conditions.

In molecule generation, models are often evaluated based on criteria like validity, uniqueness, and novelty. A tradeoff exists between validity and uniqueness: the more distinct or novel a generated molecule is, the harder it is to ensure its validity. This tradeoff is especially important in applications like antibacterial drug design, where novelty is critical to address bacterial resistance. Users can adjust priorities by tuning parameters such as the temperature in autoregressive Transformer models, emphasizing either validity or uniqueness depending on the task.
Several studies explore this tradeoff. For instance, MCMG \cite{wang2021multi} uses reinforcement learning to optimize multiple molecular properties while maintaining both validity and diversity. DiffuMol \cite{peng2024hitting} leverages a diffusion framework that provides stepwise control over multiple objectives, making it easier to generate novel molecules while optimizing specific properties. DiffuMol’s use of the Transformer’s attention mechanism helps balance diversity and specificity by focusing on key regions of the molecule that must meet certain property or scaffold requirements, while allowing variation in other parts to promote molecular diversity.

\subsection*{Sampling Algorithms}

One part of autoregressive generation process that can affect the natural plausibility of the output sequences is the sampling algorithm used to predict tokens based on the model's probability distribution. 
ProtGPT2\cite{ferruz2022protgpt2} explored various sampling strategies to generate \textit{de-novo} protein sequences, such as greedy search, beam search, and random sampling. They motivate this by noting that greedy and beam search often lead to repetitive and deterministic sequences, which do not reflect the natural variability found in proteins. On the other hand, random sampling from the top ``k'' tokens, especially when using a high value of "k" (e.g., k = 950), was found to produce more natural-like sequences by introducing variability and capturing natural amino acid propensities.

\subsection{Multi-Modal FMs}

While FMs often excel at learning from a single data modality, biological systems are inherently multi-modal, involving interactions across sequence, structure, and function at various scales. Effectively integrating these diverse data types remains an open challenge in biological modeling. Often, biological data sources are treated independently, overlooking the interactions between systems. DNA sequences are often viewed in isolation, but DNA alone does not fully describe an organism's phenotype. As Denis Noble\cite{noble2012theory,noble2024s} and others emphasize, biological processes result from interactions across molecular, cellular, physiological and organismal scales\cite{ramsden2023bioinformatics}. 

\subsection*{Combining Multiple Forms of Biological Sequences}

Multimodal approaches enhance GLMs' understanding by incorporating epigenetic and transcriptomic data. 
The gLM2 model \cite{cornman2024omg}, trained on the Open MetaGenomic corpus, uses both nucleotide sequences and corresponding amino acid sequences from coding regions, along with strand direction information, helping process gene regulation and reading frames.

Hybrid approaches that combine multiple input modalities or architectures are increasingly common in biological design. For example, MMDiff\cite{morehead2023towards} integrates both structural (using rigid body frames for rotation and translation) and sequence data to jointly generate nucleic acid-protein complexes.
RFAA\cite{krishna2024generalized} tackles the complexity of modeling biomolecular systems, including proteins, nucleic acids, small molecules, and metals. RFAA uses a three-track network architecture inspired by RoseTTAFold2 \cite{baek2023efficient} with (1) 1D sequence data for amino acids, nucleic acids, and atom types, (2) pairwise relationships like residue distances or bond types, and (3) 3D coordinate information, including atom positions and stereochemistry. RFAA also incorporates a conditional diffusion fine-tuning approach to generate binding proteins, starting from substructures like ligands.

\subsection*{Combining Sequence and Structure}

Integrating structural information with sequence data can significantly improve model performance, especially for tasks where geometry and physical constraints are crucial. A common approach is to combine molecular graphs with sequence data, providing geometric context that complements raw sequences. Protein structures—secondary, tertiary, and quaternary—also offer essential spatial information that sequence data alone may not reveal. Additional inputs can include biological environment details, target receptors, or transcriptomic data linking RNA, DNA, and amino acid sequences.

For example, models like MMDiff \cite{morehead2023towards}, which generate nucleic acid-protein complexes, use both sequence and structural data for more accurate macromolecule modeling. This is intended to help with handling Intrinsically Disordered Regions is challenging since their structure cannot be fully inferred from sequence alone, underscoring the need for both representations.
ProtMamba \cite{sgarbossa2024protmamba} enhances protein inpainting and de novo protein generation by using homologous sequences, without relying on multiple sequence alignments (MSAs). This approach captures evolutionary conservation and variability, providing valuable context for tasks like protein editing.
The Gene Ontology (GO) \cite{ashburner2000gene} also provides a structured framework for annotating biological processes, molecular functions, and cellular components. Although models like Progen2 \cite{nijkamp2023progen2} partially leverage GO terms for conditioning, the full graph structure of the ontology, which allows for rich parent-child relationships, remains underutilized for generation.

\subsection*{Incorporating Natural Language and Domain Knowledge}
One of the most appealing aspects of the FM paradigm, particularly with models like 
GPT-4, is their ability to interact via natural language, offering flexibility that is largely absent in biology-specific models. 
To bridge this gap, recent works have focused on integrating natural language with biological sequence models, enabling tasks to be performed through natural language prompts. Notable examples include ChatNT \cite{richard2024chatnt} for genomics and Nach0 \cite{livne2024nach0} and InstructBioMol \cite{zhuang2024instructbiomol} for chemical sequences.

ChatNT is a multimodal agent that combines biological sequence processing with NLP, handling DNA, RNA, and proteins through conversational input. It integrates the Nucleotide Transformer \cite{dalla2023nucleotide} with the Vicuna-7B language model using a Perceiver-based projection layer, allowing for seamless interaction in biological tasks through natural language.
Nach0 extends this concept to SMILES molecular sequences, using training data from PubMed abstracts and patents. 
It has been applied to tasks like molecular property prediction, generation, and reaction prediction. 
In a case study, Nach0 was prompted to ``Generate a random druglike small inhibitor molecule for Janus Kinase 3 (JAK3) that contains a classic kinase hinge binding motif.'' Despite lacking 3D structural information, Nach0 successfully generated eight valid molecules, with a discovery rate of 0.11\%, compared to 1.53\% from a structure-aware baseline.
InstructBioMol takes this further by training on natural language, 2D molecular graphs, protein sequences, and 3D structures. Its Motif-Guided Multimodal Feature Extraction Module allows the model to handle tasks like molecule captioning, description-based molecule and protein generation, and protein property question answering.

\section{Open Problems}

\textbf{Which Architectures for Which Biological Task:} 

Many studies introduce specific approaches (e.g., Transformers, diffusion, SSMs) for distinct applications, but there is little consensus on which architecture is best across biological tasks. Direct comparisons are challenging due to inconsistent evaluations and benchmarks. We propose exploring architectures that are underused in certain domains, such as diffusion models for textual molecular representations. Additionally, we would compare the effects of different pre-training objectives, especially since many current objectives are borrowed from NLP, where sequence behavior differs from biological data. Developing domain-specific, self-supervised objectives—both multi-modal and sequence-only—would help create better representations for biological tasks where additional modalities are unavailable.

\textbf{Innovative Architectures to handle Data Limitations:}

While FMs have shown strong performance in specific biological domains, scalability and generalization remain difficult due to the relatively small and fragmented datasets in biology. In NLP, large datasets have driven breakthroughs in models like GPT-3 \cite{brown2020language}, with scaling laws analyzed by Kaplan et al. \cite{kaplan2020scaling} and Bahri et al. \cite{bahri2024explaining}. Similar trends have been observed in biological models \cite{madani2023large, rives2021biological, frey2023neural}, but addressing data limitations may require lighter architectures or novel methods for obtaining generalizable representations without massive datasets. This demands new strategies for scalable biological models.

\textbf{Enhancing Generalization and Transfer to Low-Data Regimes:} 
Finally, we would like to explore the important application transfer learning to help CLMs generalize to classes of molecules that do not have a lot of available training data. Specifically, Quaternary Ammonium Compounds are a group of chemicals with important antibiotic properties but currently the number of available examples of molecules in this class is small. For example, we may experiment using techniques similar to Progen2's\cite{nijkamp2023progen2} antibody generation approach combining fine-tuning with conditional generation to overcome the bias from pre-training. We would like to explore which architectures best enable transfer learning in this setting and additional techniques to guide generation toward the correct region of chemical space.

\textbf{Better Integration of Biological Modalities:} 

A promising direction is the integration of various biological modalities, such as combining chemical graphs with SMILES representations in CLMs. Techniques like Adapters, used by Liu et al. \cite{liu2023molca} for molecular captioning, may also improve molecule generation. We would begin by evaluating the impact of these approaches by testing generated sequence validity and property control. Furthermore, we aim to explore pre-training models with Gene Ontology graph structures, extending Progen's use of GO-derived keywords, to enhance protein generation.

\textbf{Enabling Better Control in Generation:} 

We would like to push the boundaries of FMs' capabilities to generalize to rare or physically unlikely property combinations in the biological sequence space. This would involve studying whether the difficulty in generating these combinations is due to their absence from training data or intrinsic physical constraints. This investigation connects to Ferruz et al.'s \cite{ferruz2022protgpt2} concept of "dark" regions in protein space, and could provide insight into improving generation for uncommon combinations.

\section{Conclusion}

The field is moving ahead rapidly; a large number of the cited papers here are from preprint servers such as arXiv and BioRxiv, highlighting the need for standardization of datasets and metrics. Furthermore, looking ahead, it is crucial for research to prioritize rigorous comparative evaluations of different model architectures for biological generation, more sophisticated integration of biological modalities, and improved strategies for generalization across diverse biological systems. This is especially timely, as the United States National Academies of Sciences, Engineering, and Medicine are founding a committee on Foundation Models for Scientific Discovery and Innovation, underscoring the growing importance of this area.\cite{national_academies_fm2024} By addressing these challenges, FMs could truly revolutionize fields such as drug discovery, synthetic biology, and genetic engineering, accelerating breakthroughs and moving us closer to practical, AI-driven biological innovations that can meaningfully impact our understanding and manipulation of life.

\section*{Acknowledgements}
This work was supported in part by the National Science Foundation Grant No. 2411529 and Grant No. 2310113.


\newpage
\clearpage

\end{document}